\documentclass{article}

     \usepackage[preprint,nonatbib]{neurips_2020}

\usepackage[utf8]{inputenc} 
\usepackage[T1]{fontenc}    
\usepackage{hyperref}       
\usepackage{url}            
\usepackage[english]{babel}
\usepackage{tablefootnote}
\usepackage[utf8]{inputenc}
\usepackage{float}
\restylefloat{table}
\usepackage{booktabs}       
\usepackage{amsthm,amsmath,amsfonts}
\usepackage{mathtools}      
\usepackage{nicefrac}       
\usepackage{microtype}      
\usepackage{graphicx}
\usepackage{subfigure}
\usepackage{fourier}
\usepackage{amssymb} 
\usepackage{amsmath}

\usepackage[ruled,vlined,linesnumbered]{algorithm2e}
\usepackage[english]{babel}

\title{An efficient scheme based on graph centrality to select nodes for training for effective learning}

\author{%
  CR Sandeep$^*$, Asif Salim$^{**}$, R Sethunadh$^*$, S Sumitra$^{**}$ \\
  *Vikram Sarabhai Space Centre,\\Thiruvananthapuram, India\\
   
  **Department of  Mathematics,
  Indian Institute of Space Science and Technology,\\
  Thiruvananthapuram, India\\
   \texttt{cr\_sandeep@vssc.gov.in, asifsalim.16@res.iist.ac.in,}\\ \texttt{r\_sethunadh@vssc.gov.in, sumitra@iist.ac.in}   \\
}

\begin{document}

\maketitle

\begin{abstract}
The process of selecting points for training a machine learning model is often a challenging task. Many times, we will have a lot of data, but for training, we require the labels and labeling is often costly. So we need to select the points for training in an efficient manner so that the model trained on the points selected will be better than the ones trained on any other training set. We propose a novel method to select the nodes in graph datasets using the concept of graph centrality. Two methods are proposed - one using a smart selection strategy, where the model is required to be trained only once and another using active learning method. We have tested this idea on three popular graph datasets - Cora, Citeseer and Pubmed- and the results are found to be encouraging.
\end{abstract}
\section{Introduction}
Consider the problem of classifying the vertices in a graph, $G=(V,E)$, where $V$ is the set of vertices with $|V|=N$,$N$ being the number of nodes and $E$ is the edge set. The usual way adopted in Graph Semi-Supervised Learning (SSL) is to select a few number of nodes upfront randomly as the nodes for which we seek the labels from an Oracle and use these nodes for training the model. There are number of attempts to effectively perform this task, the popular one being the approach of using Graph Convolutional Networks (GCN) as proposed by Kipf et.al.$\cite{r15}$ The nodes which we select for labeling may not be the most ideal or informative points for the model to learn, so if we select the points in a better way, the model also can be made better. \\

Another popular idea for selecting the points for training is Active Learning, where the model is initially given a few points for training and from the trained model, new points are selected for training based on some form of acquisition function, which uses either the model uncertaininty or the ability to cover the data distribution as a measure to rank the points. The penalty involved in such a method is that we need to re-train the model every time after we select the points. \\

To make a better selection of nodes for training, we propose to use the concept of Graph Centrality. Graph Centrality is a concept from graph theory, where the centrality of a node is defined as the `importance' of that node in the graphical structure. This importance can be defined in many ways, degree centrality, closeness centrality, shortest path centrality, page rank centrality etc being the most popular measures of centrality. \\

Our attempt is to select the nodes using the centrality, so that the nodes which are selected for training will be more `important' and hence more `informative' for the model to learn. We propose a simple but efficient method to select the points based on centrality and the experiments have shown that this selection indeed results in models which are better than the ones which select the points randomly.\\

Two methods are proposed - First Smart Selection, where we select the training points upfront using graph centrality on the existing graph structure and carry out the training only once. Second, active learning, where we apply graph centrality on the features learned using a neural network and then selecting points for training in batches. To the best of our knowledge, this is the first attempt to make use of graph centrality to select the points efficiently for training.\\

Our contribution in this paper are summarized as below:
\begin{enumerate}
	\item Using the concept of graph centrality to select the training points
	\item Using graph centrality combined with graph convolutional networks(GCN) to select the training points in a smart way so that the model trained on the points selected gives better results
	\item Applying the same concept in the active learning set up 
\end{enumerate}
The rest of the paper is organized in the following manner. Section 2 describes the similar works carried out so far. Section 3 gives the background on graph centrality and graph convolutional networks. The proposed scheme is explained in Section 4. Results and the discussions on the results are given as Section 5 and Section 6 is the conclusion part.
\section{Related Works}
\textbf{Semi-Supervised Classification on Graphs}: The spectral graph convolution neural networks was introduced by Bruna et al.\cite{r21} where the Graph Fourier Transform on the nodes of the graph was utilized. The convolution is defined on the space spanned by the eigenvectors of the graph Laplacian. In the work of Defferrard et al.\cite{r17}, convolution filter is defined as the linear combination of the powers of Laplacian. Kipf et al. \cite{r15} modified this approach by using the first order approximation of the convolution defined using the graph Laplacian. In the work of Xu et al.\cite{r22}, a method GraphHeat is put forward, where a negative exponential on the eigenvalues of the Laplacian is used, which penalizes the higher frequency components. There were a number of attempts which tried to improve the GCN. Some of the important ones are Graph Attention Networks(Velickovic et al.\cite{r24}), which introduced the concept of attention to the graph convolutional networks, Conf GCN by Vashishth et al. \cite{r18}, which estimated the label score along with their confidences jointly in a GCN setting.\\ 

\textbf{Active Learning}:Active Learning algorithm has to learn from a small amount of data and also come up with ways to represent the uncertainities in the model effectively. This restricts the models that are available for active learning especially in deep learning scenarios. The earlier works in this direction were mostly based on low-dimensional problems such as the one suggested in \cite{r2}. New algorithms were developed to process the higher dimensional data in the deep learning era and especially with the arrival of Convolutional Neural Networks (CNN).\\

There are new approaches like the one mentioned in the work of Yoo et al.\cite{r4}, where given a training data point $x$, target model gives $\hat{y}=\Theta_{target}(x)$ and a predicted loss through the loss prediction module as $\hat{l}=\Theta_{loss}(h)$, where $h$ is the feature set of $x$ extracted from the several hidden layers of target model. Since annotation for $x$, $y$ is known, the target loss is computed as $l=L_{target}(y,\hat{y})$. Since this loss $l$ is a ground-truth target of $h$ for the loss prediction module, we can also compute the loss for the loss prediction module as $L_{loss}(l,\hat{l})$.\\

The method prescribed in the work of Sener et al.\cite{r7}, where the idea is to allow thes training set to capture the diversity in the dataset. To find the data that isn’t represented well yet by the training set, we need to find the “Core Set” at every step. The core set is defined as a batch of points such that when added to the training set, the distance between a point in the unlabeled pool and it’s closest point in the training set, will be minimized. \\

Most of the Active Learning algorithms acquisition function relies on the uncertainity of model predictions. But deep learning rarely represent this uncertainity. There are specialized models such as Deep Bayesian CNNs \cite{r13}, which has formed the base line for the paper by Gal et al.\cite{r1}. To predict the model uncertainities for Active Learning scenario, stochastic regularization techniques such as drop-out is used. Inference is done by training a model with dropout before every weight layer, and by performing dropout at test time as well to sample from the approximate posterior. The idea is to try and get the uncertainity measure of the neural network through the drop-out.\\

\indent Another novel idea, which is discussed in the work of Kushnir et al.\cite{r20}, is a diffusion based approach. In this, the data points are represented as a graph whose node representations are taken from the penultimate layer of a neural network. A graph is constructed from K-Nearest Neighbours and an algorithm is used to find the most informative points for annotation.
\section{Background-Graph Centrality \cite{r25}}
In Graph Theory, Centrality measures try to identify the most important vertices in a graph.  Graph Centrality is defined as a real-valued function on the vertices of a graph, where the values produced are expected to provide a ranking which identifies the most important nodes. he definition of 'importance' may vary according to the context. The popular measures of 'importance' are Degree, Betweenness, Closeness etc
\subsection{Degree Centrality}
This is one of the basic measures of centrality. It is equal to the number of immediate neighbors of a node. That is Degree is simply the number of nodes at distance one. Thus the more neighbors a node have the more central and highly connected that node is. We assume that the higher degree nodes have more influence on the graph. Though simple, degree is often a highly effective measure of the influence of a node.
\subsection{Closeness Centrality}
This measures the 'closeness' of a node to all other nodes. The farness of a node $\mathnormal{v}$ is defined as the sum of its distances to all other nodes and closeness is the inverse of farness.
\begin{equation*}
closeness(\mathnormal{v})=\frac{1}{\sum_{\mathnormal{i}\neq \mathnormal{v}}d_{\mathnormal{v}\mathnormal{i}} }
\end{equation*}
The more central a node is, the lower its total distance to all other nodes.
\subsection{Betweenness Centrality}
 Betweenness centrality quantifies the number of times a node acts as a bridge along the shortest path between two other nodes. The vertices that have a high probability to occur on a randomly chosen shortest path between two randomly chosen vertices have a high betweenness
\begin{equation*}
Betweenness(\mathnormal{v})=\sum_{\mathnormal{s}\neq \mathnormal{t}\neq \mathnormal{v} \in V}\frac{\sigma_{\mathnormal{st}}(\mathnormal{v})} {\sigma_{\mathnormal{st}}}
\end{equation*}
Where $\sigma_{\mathnormal{st}}$ is the total number of shortest paths from $\mathnormal{s}$ to $\mathnormal{t}$ and $\sigma_{\mathnormal{st}}(\mathnormal{v})$ is the number of those shortest paths which pass through $\mathnormal{v}$
\subsection{Pagerank Centrality}
Here Google PageRank algorithm is used to rank the nodes. We define a column stochastic matrix 
\begin{equation*}
N=AD^{-1}
\end{equation*}
where $A$ is the Adjascency matrix of the graph and $D$ is the diagonal matrix, with $D_{ii}=\sum_j A_{ij}$. We need to ensure that the link matrix is a positive column stochastic matrix. So the link matrix to find the PageRank is given by,
\begin{equation*}
M=(1-\alpha)N+\alpha S
\end{equation*}
where $\alpha \in [0,1]$ and $S=\frac{1}{N}\mathnormal{I}$, $N$, being the number of nodes and $\mathnormal{I}$ represent the Identity Matrix. $M$ is a positive column stochastic matrix which will have only one L.I eigen vector corresponding to eignevalue=1.\\

PageRank centrality of the node $\mathnormal{i}$ is the rank of $\mathnormal{x}_\mathnormal{i}$ in the eigenvector $\mathnormal{x}=M\mathnormal{x}$
\subsection{Voterank Centrality}
The paper by Zhang et al.\cite{r23}, presents a centrality measure called Voterank Centrality which is used to identify a set of decentralized spreaders with the best spreading ability. This is a paper in the area of influence maximization problem in graph theory. As influential nodes have strong ability to affect other nodes, selecting top-ranked influential nodes as source spreaders is a common strategy. The key idea here is that if one node A has the supported another node B, then the support strength of A to other nodes will decrease. Under this approach, a Vote based algorithm to identify the most influential nodes is put forward.\\

In this algorithm, each node $\mathnormal{u}$, is attached with a tuple $(s_{\mathnormal{u}},va_{\mathnormal{u}})$, where $s_{\mathnormal{u}}$ is the voting score, which is the number of votes the node $\mathnormal{u}$ has obtained from its neighbors and $va_{\mathnormal{u}}$ is the value of vote which $\mathnormal{u}$ can give to its neighbors. \\

The algorithm is as follows:
\begin{itemize}
	\item \textbf{Step-1} - Initialize - For all nodes, $(s_{\mathnormal{u}},va_{\mathnormal{u}})=(0,1)$
	\item \textbf{Step-2} - Vote. The nodes vote for their neighbors. After this, voting score of each node will be calculated. Voting score of a node already selected as a leading node in an earlier round will be set to zero so as to avoid electing it again. 
	\item \textbf{Step-3} - Select the node which gets the maximum votes. This node will not participate in the subsequent votings as its voting ability also will be set to zero
	\item \textbf{Step-4} - Update the voting abilities - Weaken the voting ability of the nodes which has voted for the node which got selected in the earlier round
	\item \textbf{Step-5} - Repeat the Steps-2 to 4 until we select the $\mathnormal{r}$ number of influential nodes 
\end{itemize}
\section{Background-Graph Convolutional Networks}\
Let $G=(V,E)$ represents a graph with $V$ being the set of vertices and $E$ the set of edges between the vertices. The adjacency matrix is defined as $W$, where $W_{ij}$ denotes the weight of the edge $(i,j)$ and 0 if there is no edge between two vertices. The degree matrix $D$ is defined as a diagonal matrix with $D_{ii}=\sum_j W_{ij}$. If $G$ is an un-weighted graph, the adjacency matrix entries are defined as $W_{ij}=1$, if $i \sim j$. For undirected graphs, $W$ will be a symmetric matrix.\\

The Laplacian of the graph is defined as $L=D-W$ and the normalized Laplacian is $ \tilde{L}=D^{-\frac{1}{2}}LD^{-\frac{1}{2}}=I-D^{-\frac{1}{2}}WD^{-\frac{1}{2}}$. The $\tilde{L}$ is a real symmetric positive semi definite matrix and has a complete orthonormal eigenvectors $\{u_l\}_{\mathnormal{l}=1}^{\mathnormal{l}=N}$, and the associated ordered real non negative eigenvalues $\{\lambda_l\}_{\mathnormal{l}=1}^{\mathnormal{l}=N}$. \\

Graph Fourier transform is a transformation where we decompose the Laplacian matrix into its eigenvalues and eigenvectors. It is a process of projecting a signal $x$ defined on the graph to the space spanned by the eigenvectors of the graph Laplacian. The eigenvalues of $\tilde{L}$ are called the frequencies of the graph and the eigenvectors are called the Fourier modes for the graph. We could define the eigen decomposition of $\tilde{L}$ as $\tilde{L}=U\Lambda U^T$, where $U$ is the matrix of eigenvectors and $\Lambda$ is the diagonal matrix of eigenvalues. The Graph Fourier transform of a signal $\mathnormal{x}$, given by, $\mathnormal{F}(x)=\hat{\mathnormal{x}}=U^T \mathnormal{x}$. The Inverse Graph Fourier Transform is defined as,$\mathnormal{F}^{-1}(\hat{\mathnormal{x}})=\mathnormal{x}=U\hat{\mathnormal{x}}$, where $\hat{\mathnormal{x}}$ is the resultant signal from the Fourier Transform.
\subsection{Spectral Graph Convolutional Networks}
The spectral convolution on a graph can be defined as the defined as the convolution of
a signal $x \in \mathbb{R}^N$ with a filter $g \in \mathbb{R}^N$. This can be written in the Fourier domain as the multiplication of the Fourier transforms. So,$g \circledast \mathnormal{x}  =\mathnormal{F}^{-1}(\mathnormal{F}(g) \odot \mathnormal{F}(x) )= U(U^Tg \odot U^Tx)$. \\

Considering $g$ being parametrized by $\theta$, and $g_{\theta}$ can be considered as function of eigenvalues of $\tilde{L},g_{\theta}(\Lambda) \in \mathbb{R}^{N\ X\ N}$ (where $\Lambda$ is the diagonal matrix of eigenvalues of $\tilde{L}$), we get
$ g_{\theta} \circledast \mathnormal{x} = Ug_{\theta}(\Lambda)U^T \mathnormal{x}$.\\
The computation of the eigen decomposition and the matrix multiplications are costly especially for large graphs. Hence the method suggested in \cite{r14} is used, where $g_{\theta}(\Lambda)$ is approximated by truncated expansion in terms of Chebyshev polynomials upto $K^{th}$ order. So the expression reduces to $K^{th}$ order polynomial in the Laplacian, ie: the nodes that are $K$-steps away from a central node only are considered.\\

So the spectral convolution can be approximated \cite{r15} as  $ g_{\theta} \circledast \mathnormal{x} \approx  U \left ( \sum_{k=0}^{K} \theta_k T_k(\Lambda) \right ) U^T \mathnormal{x} $
where $T_k(\mathnormal{x})$ is the $k^{th}$ order Chebyshev Polynomial. Since $\tilde{L}=U\Lambda U^T$ and $(U\Lambda U^T)^k=U\Lambda^k U^T$, we get $g_{\theta} \circledast \mathnormal{x} \approx  \sum_{k=0}^{K} \theta_k T_k(\tilde{L}) \mathnormal{x}$. Limiting the $K=1$, ie: as a linear function on the graph Laplacian, also assuming $\theta_0=-\theta_1=\theta$, to reduce the number of parameters and then applying the renormalization trick, where we add the self loop connection to ensure the training stability, we get
$ g_{\theta} \circledast \mathnormal{x} \approx \theta(\tilde{D}^{-\frac{1}{2}}\tilde{W}\tilde{D}^{-\frac{1}{2}})\mathnormal{x}$, where $\tilde{W}=W+I$ and $\tilde{D}_{ii}=\sum_j \tilde{W}_{ij}$
\section{Proposed Scheme}
\subsection{Problem Statement}
Given a graph $G=(V,E)$, where $V$ is the set of nodes and $|V|=N$, the number of nodes and $E$ is the edge set defined on the vertices of $G$. Let the training points for the Semi-Supervised Learning, be $X_{train}$ and let $N_{train}$ be the number of points that can be chosen for training. For these $N_{train}$, we need to get the labels and using these labels we train the model. The model can be any neural network model, but for this experiment, we define the model defined by Kipf et al. \cite{r15} as our model. \\

The problem is to find the optimum $X_{train}$ using the graph centrality measures, so that the model trained on the points selected will be better than the models trained on any other points.
\subsection{Node Selection for Training using Graph Centrality}
Since the Graph Centrality tries to rank the nodes based on their importance/influence, can we use the concept of centrality to select the points for training. For the Graph datasets, the graph structure is already defined and we could directly apply centrality measures to find out the important nodes. From the graph structure we could extract the adjacency matrix and using the adjacency matrix, we could compute the importance score for each node based on the centrality. From this, we could select the nodes having high centrality scores and use them for training. 
\subsection{Selecting all nodes at once for Training}
This is the most simple method to select the points for training. Here, after finding the adjacency matrix $W$, we apply different centrality measures. The top $N_{train}$ points from the centrality list are selected as the training points. The training is done only once. 
\subsection{Smart Selection}
The points which are very high on the centrality measure, may have neighbours who also tend to be more central. This may lead to a scenario where the points selected on centrality may be more clustered and hence may not represent the diversity present in the dataset. To avoid this, a different, but simple approach is introduced. We call this \textit{`Smart Selection'} which is given in the following algorithm.
\begin{itemize}
	\item \textbf{Step-1} - Make the adjacency matrix, $W$\\
	\item \textbf{Step-2} - Using the adjacency matrix $W$, find the centrality of nodes\\
	\item \textbf{Step-3} - From the centrality list, choose the Top-10 nodes and assign it as training points\\
	\item \textbf{Step-4} - Remove these nodes from the graph and find the centrality again\\
	\item \textbf{Step-5} - Repeat \textbf{Step-3} and \textbf{Step-4} till we reach the labelling budget($N_{train}$)\\
	\item \textbf{Step-6} - Once the labelling points are selected, do the training with these selected points as the training set\\
	\item \textbf{Step-7} - Apply the model on the test set
\end{itemize}
The intuition here is that, when we select only top-10 or top-5 points based on centrality and then by removing them we ensure that the training points are not concentrated on one particular part of the graph. When the top points are removed, and we find the centrality, the new top central points tend to cover the parts of the graph which were not covered initially. By continuing this, till we reach the labelling budget, the graph structure is well covered and hence the chance of mapping the distribution effectively is more.\\

For Voterank centrality though, the process of diminishing the effect of the neighbours of a highly central node is done within the centrality algorithm itself. So for Voterank centrality, we could select the top points all at once.
\subsection{Active Learning using Graph Centrality}
For Active Learning, we need to select the data points/nodes which are more informative so that the model trained on those points will be better than the model trained on other points. Since the Graph Centrality tries to rank the nodes based on their importance/influence, the question is whether we can use the concept of centrality to select the points for Active Learning. Here instead of using the graph structure as it is, the attempt is to utilize the features extracted using a neural network. \\

For this, we start with a small number of training points for which the labels are sought and using them, we train a GCN and the feature reprsentation from the penultimate layer of the neural network is taken. The final layer being the softmax classification layer, penultimate layer gives more of the features learned. Then using the features learned, we could construct another graph by applying nearest neighbours on the feature space. On the newly constructed graph, we can apply the centrality measures and select the important points based on the centrality and for the top most central points, labels can be sought from an oracle and re-train the model. This process can be repeated till we reach the labelling budget and the effectiveness of the method can be studied.\\

Since the process of finding the centrality is applied on a graph which is created from the neural network representations, the idea can be applied to regular non-graph datasets as well. 
\subsection{Implementation}
\subsubsection{Datasets}
Three citation datasets - Cora, Citeseer and Pubmed, are used in the experiments. The nodes are documents and the edges are citation links. The Cora dataset consists of 2708 scientific publications classified into one of seven classes. The CiteSeer dataset consists of 3312 scientific publications classified into one of six classes. The Pubmed dataset consists of 19717 scientific publications from PubMed database pertaining to diabetes classified into one of three classes. The details of each of the dataset is listed in Table-\ref{tab-1}. 
\begin{table}[H]
\begin{center}
	\caption{Datasets Details}
	\begin{tabular}{||l l l l l||} 
		\hline
		\textbf{Dataset} & \textbf{Nodes} & \textbf{Edges} & \textbf{Classes} & \textbf{Features}\\[0.5ex]
		\hline
		Cora     & 2,708  & 5,429  & 7 & 1,433 \\
		Citeseer & 3,327  & 4,732  & 6 & 3,703 \\
		Pubmed   & 19,717 & 44,338 & 3 & 500\\
		\hline
	\end{tabular}
	\label{tab-1}
\end{center}
\end{table}
\subsubsection{Baseline}
To compare the performance of the selection of training points in the methods we have introduced, we use GCN (Kpif et al.\cite{r15}). In this semi-supervised learning setup, 140 training points are chosen at random. 1000 points are kept aside for testing. The task is to predict the labels on the test set. A 2-layer GCN is used for the classification purpose.
\begin{equation}
Z=f(X,W)=softmax(\hat{W}ReLU(\hat{W}X\Theta^{(0)})\Theta^{(1)})
\end{equation}
where $\Theta^{(0)}$ is the Input-to-Hidden weight matrix, $\Theta^{(1)}$ is the Hidden-to-Output weight matrix. The Loss used is the negative log-likelihood loss.\\

For all three datasets, 140 points are the labelling budget for training. For baseline method, the first 140 points are chosen for training. For the other methods, the training nodes are selected from a pool of 500 nodes. The test set consists of 1000 nodes, which are used exclusively for testing only.The task is to train a neural network such that the network should be able to correctly predict the class for the testing nodes. The performance measure used is Accuracy of predictions.\\

\subsubsection{Selecting the Training Set using Centrality}
The labelling budget for all three datasets is fixed as 140 points as done in \cite{r15}. The training points are chosen from a set of 500 points using the centrality measures. Different centrality measures are applied on the datasets. Two methods were attempted, the first one where all 140 points are selected after applying the centrality and the second one where points are selected using the Smart Selection algorithm mentioned above.\\

For the active learning purpose, we start with 10 points initially, and train a GCN. Then from the model learned, another graph is constructed using sklearn's nearest neighbors. On this graph, different centrality measures are applied and the top-10 points are chosen for labelling in each iteration. Using the newly added labelled points, network is re-trained and centrality is applied again. This process is repeated until we reach the labelling budget. 
\section{Results and Discussion}
\subsection{Results}
The results of the method where we select all the points at once, using the smart selection without using the active learning and when we use active learning is shown below in 3 tables for 3 datasets. The methods are ranked based on the statistical t-test to check the significance of the results.
\begin{table}[H]
\begin{center}
	\caption{Results}
	\begin{tabular}{||c | c | l | l | l||} 
		\hline
		\textbf{Rank} & \textbf{Method} & \textbf{Cora} & \textbf{Citeseer} & \textbf{Pubmed} \\ [0.5ex] 
		\hline\hline
		1 &\textbf{SS-Degree} &  \textbf{84.23$\pm$0.05} &  \textbf{72.78$\pm$0.35} &  \textbf{82.69$\pm$0.23}\\
		\hline
		2 & SS-Closeness & 83.53$\pm$0.024 & 71.59$\pm$0.36 & 79.5$\pm$0.34\\
		\hline
		2 & SS-Betweeness & 83.3$\pm$0.06 & 72.19$\pm$0.21 & 80.1$\pm$0.21\\
		\hline
		3 & AL-Voterank & 82.5$\pm$0.004 & 72.8$\pm$0.16 & 79.6$\pm$0.27\\
		\hline
		4 & AL-Degree & 82.7$\pm$0.014 & 70.4$\pm$0.34 &  79.8$\pm$0.31 \\
		\hline
		5 & AL-Closeness & 82.4$\pm$0.06 & 67.9$\pm$0.21 &  79.5 $\pm0.34$\\
		\hline
		5 & SS-Voterank & 82.34$\pm$0.07 & 70.3$\pm$0.23 & 78.82$\pm$0.35\\
		\hline
		5 & AL-Pagerank & 80.5$\pm$0.005 & 72.1$\pm$0.32 & 78.5 $\pm$0.36\\
		\hline
		6 & GCN \cite{r15} & 81.5$\pm$0.63 & 70.3$\pm$0.54 & 78.48$\pm$0.58\\
		\hline
		6 & SS-Pagerank & 81.03$\pm$0.064 & 70.59$\pm$0.30 & 78.53$\pm$0.36\\
		\hline
		7& AL-Betweeness & 78.4$\pm$0.06 & 67.2$\pm$0.24 & 76.3$\pm$0.44\\
		\hline
	\end{tabular}
	\label{tab-2}
\end{center}
\end{table}
Note: SS-Smart Selection. AL-Active Learning\\
The rank represent the average rank for that method based on the statistical t-test. Same rank for different methods tells that statistically there is no significant difference among those methods.\\
	
The t-SNE Visualization of the GCN embeddings of the points and the training points selected for the Cora datset is shown below. The 16 dimensional GCN output is reduced to 2 dimensional representation using t-SNE.
\begin{figure}[H]
	\centering
	\includegraphics[scale=0.65]{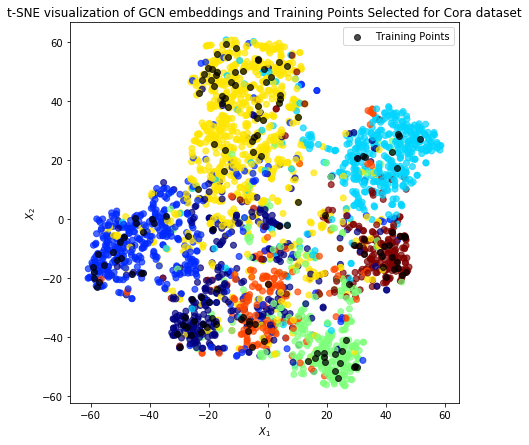}
\end{figure}
\subsection{Time Complexity}
The processing involved in Smart Selection compared to the baseline GCN is the process of applying graph centrality, selecting the top nodes, deleting those nodes and repeating the process till we reach the labelling budget. So the major contributer to time complexity is the time required to compute the graph centrality. This differs with the centrality measure used.\\

In the case of Active Learning, the process is a little more complex. In this case, we train a neural network with a small number of nodes first, and then using the features learned, a graph is contructed and centrality is applied. Then from the centrality measures, we select more points for training and the whole process is repeated. So each cycle of active learning involves training, construction of graph from the features learned and computing centrality. All these three steps contribute to time complexity.\\

A comparison of time complexity for both the methods is carried out for Cora dataset. The experiment is carried out in CPU for all three cases. It will be significantly lower if GPU is used. The timings are presented in Table-\ref{tab-3}.
\begin{table}[H]
	\begin{center}
		\caption{Time Complexity for Cora Dataset}
		\noindent\makebox[\textwidth]{%
		\begin{tabular}{||c| c c | c | c c c | c||} 
			\hline
			\textbf{Centrality Used} & \multicolumn{3}{|c|}{\textbf{Smart Selection}} & \multicolumn{4}{|c||}{\textbf{Active Learning}} \\
			\hline
			\textbf{ } & \multicolumn{2}{c|}{Time taken for} & \textbf{ } & \multicolumn{3}{c|}{Average Time taken for} & \textbf{}\\
			\textbf{ } & \textbf{Centrality} & \textbf{Training} & \textbf{Total Time} &  \textbf{Construct Graph} & \textbf{Centrality} & \textbf{Training}  & \textbf{Total Time} \\
			\hline
			Degree & 0.779s & 5.92s & 6.73s & 0.434s & 0.002s & 5.62s  & 108.10s \\
			\hline
			Closeness & 231.4s & 5.1s & 236.53s & 0.416s & 31.72s & 5.46s & 549.36s\\
			\hline
			Betweenness & 697.21s & 6.71s & 703.95s & 0.402s & 110.62s & 6.08s & 1662.57s\\
			\hline
			Pagerank & 8.08s & 6.77s & 14.88s & 0.463s & 1.58s & 5.96s & 134.86s \\
			\hline
			Voterank & 22.45s & 6.77s & 29.26s & 0.412s & 114.88s & 5.36s & 1712.61s\\
			\hline \hline
		\end{tabular}}
		\label{tab-3}
	\end{center}
\end{table}
The training time shown is the total time taken for 500 epochs of training. The time to construct graph, select nodes and training shown in the case of active learning is the average time taken for the 14 active learning cycles, each cycle selecting 10 points. Total time taken in both cases is the time taken after reading and loading the data from the files till the end of testing.
\subsection{Discussion}
We could see that by applying a smart selection method using degree centrality, there is siginificant improvement in accuracy for all three datasets. The important points is that this simple centrality measure is the fastest to compute and is most effective. Closeness centrality also gives good results although not as much improvement as with degree-centrality. Betweenness centrality also improves the performance. But the computation cost involved in closeness and betweenness centralities is much more compared to degree centrality. This is shown in Table-3 \\

The performance for the active learning set up, does improve for degree and vote rank centralities. But not as good as the one with degree centrality and smart selection. But this could be attempted in the case of non-graphical data, where we can construct a graph from the features extracted using the neural network.\\

From the t-SNE visualization of the GCN embeddings, we could see that the training points, which are shown in black colour, are well spread across all classes. That is the training points for Cora dataset are distributed effciently. Similar observation is made for the other two datasets as well.\\

We could also observe that the time to compute degree centrality is the lowest for both Smart Selection and Active Learning. The time to compute the centrality gets reduced in active learning, but this gets nullified because of re-training cycles.\\

The major point to note is that unlike in the active learning set up, where model is trained each time after selection, for the smart selection, no training is done until we select all the points upfront. Hence there is no re-training penalty is involved. So the total time taken for smart selection is considerably low compared to active learning. The vanilla GCN, where training points are selected as the first 140 points, takes 5.96s, compared to 6.73s for the method of Smart Selection with Degree centrality. So Smart selection with degree centrality comes up with a better performance with only slight increase in the time complexity. But for other centrality measures and active learning, the time complexity is higher than that of vanilla GCN.
\section{Conclusion}
In this paper, we present a way to select training points for a dataset using the concept of graph centrality. The concept is used to select the points for training using a smart selection where we do the training only once after selecting the required number of points using different centralities. The method was also applied in the active learning set up, where we select a batch of points, do the training and from the model output again select the next batch of points for training. It is seen that, when the smart selection strategy is applied using the most basic centrality-degree centrality, the results are really promising. The active learning set up also improves the performance for degree, vote rank and closeness centrality, but the computational cost involved is more. However for non-graph datasets, active learning strategy can be applied. The future work involves applying the method on tabular datasets to see whether the selection strategy yields better results for them. A similar strategy can be applied for regression tasks also.
 \bibliographystyle{ieeetr}
 \bibliography{Paper-GraphCentrality} 

\begin{thebibliography}{10}

\bibitem{r15}
T.~N. Kipf and M.~Welling, ``Semi-supervised classification with graph
  convolutional networks,'' in {\em 5th International Conference on Learning
  Representations, {ICLR} 2017, Toulon, France, April 24-26, 2017, Conference
  Track Proceedings}, OpenReview.net, 2017.

\bibitem{r21}
J.~Bruna, W.~Zaremba, A.~Szlam, and Y.~LeCun, ``Spectral networks and locally
  connected networks on graphs,'' {\em arXiv preprint arXiv:1312.6203}, 2013.

\bibitem{r17}
M.~Defferrard, X.~Bresson, and P.~Vandergheynst, ``Convolutional neural
  networks on graphs with fast localized spectral filtering,'' in {\em Advances
  in neural information processing systems}, pp.~3844--3852, 2016.

\bibitem{r22}
B.~Xu, H.~Shen, Q.~Cao, K.~Cen, and X.~Cheng, ``Graph convolutional networks
  using heat kernel for semi-supervised learning,'' in {\em Proceedings of the
  Twenty-Eighth International Joint Conference on Artificial Intelligence,
  {IJCAI-19}}, pp.~1928--1934, International Joint Conferences on Artificial
  Intelligence Organization, 7 2019.

\bibitem{r24}
P.~Veličković, G.~Cucurull, A.~Casanova, A.~Romero, P.~Liò, and Y.~Bengio,
  ``Graph attention networks,'' in {\em International Conference on Learning
  Representations}, 2018.

\bibitem{r18}
S.~Vashishth, P.~Yadav, M.~Bhandari, and P.~Talukdar, ``Confidence-based graph
  convolutional networks for semi-supervised learning,'' in {\em Proceedings of
  Machine Learning Research} (K.~Chaudhuri and M.~Sugiyama, eds.), vol.~89 of
  {\em Proceedings of Machine Learning Research}, pp.~1792--1801, PMLR, 16--18
  Apr 2019.

\bibitem{r2}
S.~Tong, {\em Active learning: theory and applications}, vol.~1.
\newblock Stanford University USA, 2001.

\bibitem{r4}
D.~Yoo and I.~S. Kweon, ``Learning loss for active learning,'' in {\em
  Proceedings of the IEEE Conference on Computer Vision and Pattern
  Recognition}, pp.~93--102, 2019.

\bibitem{r7}
O.~Sener and S.~Savarese, ``Active learning for convolutional neural networks:
  A core-set approach.,'' in {\em ICLR}, OpenReview.net, 2018.

\bibitem{r13}
Y.~Gal and Z.~Ghahramani, ``Bayesian convolutional neural networks with
  bernoulli approximate variational inference,'' {\em ICLR workshop track,
  2016a}, 2016.

\bibitem{r1}
Y.~Gal, R.~Islam, and Z.~Ghahramani, ``Deep bayesian active learning with image
  data,'' in {\em ICML}, 2017.

\bibitem{r20}
D.~Kushnir and L.~Venturi, ``Diffusion-based deep active learning,'' {\em arXiv
  preprint arXiv:2003.10339}, 2020.

\bibitem{r25}
S.~P. Borgatti and M.~G. Everett, ``A graph-theoretic perspective on
  centrality,'' {\em Social networks}, vol.~28, no.~4, pp.~466--484, 2006.

\bibitem{r23}
J.-X. Zhang, D.-B. Chen, Q.~Dong, and Z.-D. Zhao, ``Identifying a set of
  influential spreaders in complex networks,'' {\em Scientific reports},
  vol.~6, p.~27823, 2016.

\bibitem{r14}
D.~K. Hammond, P.~Vandergheynst, and R.~Gribonval, ``Wavelets on graphs via
  spectral graph theory,'' {\em Applied and Computational Harmonic Analysis},
  vol.~30, no.~2, pp.~129 -- 150, 2011.

\end{thebibliography}
\end{document}